\documentclass{article}
\usepackage{authblk}
\usepackage{hyperref}
\usepackage{enumitem}
\usepackage{booktabs}
\usepackage{tablefootnote}
\usepackage{multirow}
\usepackage{geometry}
\usepackage{graphicx}
\usepackage{float}
\usepackage{ulem}
\usepackage{makecell}
\usepackage[utf8]{inputenc}

\title{An Application of Large Language Models to Coding Negotiation Transcripts}
\author[a1]{Ray Friedman}
\author[a21]{Jaewoo Cho}
\author[b13]{Jeanne Brett}
\author[a21]{Xuhui Zhan}
\author[a2]{Ningyu Han}
\author[a2]{Sriram Kannan}
\author[a2]{Yingxiang Ma}
\author[a2]{Jesse Spencer-Smith}
\author[c3]{Elisabeth Jäckel}
\author[c3]{Alfred Zerres}
\author[a56]{Madison Hooper}
\author[a6]{Katie Babbit}
\author[a7]{Manish Acharya}
\author[ii4]{Wendi Adair}
\author[g2]{Soroush Aslani}
\author[pp4]{Tayfun Aykaç}
\author[d4]{Chris Bauman}
\author[ff4]{Rebecca Bennett}
\author[x4]{Garrett Brady}
\author[o4]{Peggy Briggs}
\author[e4]{Cheryl Dowie}
\author[4]{Chase Eck}
\author[f4]{Igmar Geiger}
\author[pp4]{Frank Jacob}
\author[cc4]{Molly Kern}
\author[bb4]{Sujin Lee}
\author[u4]{Leigh Anne Liu}
\author[kk4]{Wu Liu}
\author[v4]{Jeffrey Loewenstein}
\author[p4]{Anne Lytle}
\author[w4]{Li Ma}
\author[s4]{Michel Mann}
\author[nn4]{Alexandra Mislin}
\author[n4]{Tyree Mitchell}
\author[pp4]{Hannah Martensen née Nagler}
\author[t4]{Amit Nandkeolyar}
\author[z4]{Mara Olekalns}
\author[h4]{Elena Paliakova}
\author[jj4]{Jennifer Parlamis}
\author[i4]{Jason Pierce}
\author[gg4]{Nancy Pierce}
\author[mm4]{Robin Pinkley}
\author[pp4]{Nathalie Prime}
\author[h4]{Jimena Ramirez-Marin}
\author[q4]{Kevin Rockmann}
\author[o4]{William Ross}
\author[ee4]{Zhaleh Semnani-Azad}
\author[r4]{Juliana Schroeder}
\author[z4]{Philip Smith}
\author[pp4]{Elena Stimmer}
\author[j4]{Roderick Swaab}
\author[hh4]{Leigh Thompson}
\author[k4]{Cathy Tinsley}
\author[l4]{Ece Tuncel}
\author[y4]{Laurie Weingart}
\author[m4]{Robert Wilken}
\author[dd4]{JingJing Yao}
\author[w4]{Zhi-Xue Zhang}

\affil[1]{Project leads.}
\affil[2]{Data scientists who built the AI model.}
\affil[3]{Negotiation coding experts.}
\affil[4]{Data contributors.}
\affil[5]{Statistical analyst.}
\affil[6]{Coders.}
\affil[7]{Website designer.}
\affil[a]{Vanderbilt University}
\affil[b]{Negotiation and Team Resources (NTR)}
\affil[c]{University of Amsterdam}
\affil[d]{University of California, Irvine}
\affil[e]{University of Houston}
\affil[f]{Aalen University}
\affil[g]{University of Wisconsin, Whitewater}
\affil[h]{IESEG School of Management}
\affil[i]{UNC, Greensboro}
\affil[j]{INSEAD}
\affil[k]{Georgetown University}
\affil[l]{Webster University}
\affil[m]{ESPC Business School}
\affil[n]{Louisiana State University}
\affil[o]{University of Wisconsin, LaCross}
\affil[p]{Anne Lytle \& Associates}
\affil[q]{George Mason University}
\affil[r]{University of California, Berkley}
\affil[s]{Leuphana University}
\affil[t]{Indian Institute of Management Ahmedabad}
\affil[u]{Georgia State University}
\affil[v]{University of Illinois}
\affil[w]{Guanghua School of Management Peking University}
\affil[x]{Bocconi University}
\affil[y]{Carnegie Mellon}
\affil[z]{University of Melbourne}
\affil[bb]{KAIST}
\affil[cc]{City University of New York}
\affil[dd]{IESEG School of Management}
\affil[ee]{California State University, Northridge}
\affil[ff]{University of Central Florida}
\affil[gg]{University of North Carolina at Greensboro}
\affil[hh]{Norhwestern University}
\affil[ii]{University of Waterloo}
\affil[jj]{University of San Fancisco}
\affil[kk]{Hong Kong Polytechnic University}
\affil[mm]{Southern Methodist University}
\affil[nn]{American University}
\affil[pp]{ESCP Business School}

\begin{document}
\newcolumntype{L}[1]{>{\raggedright\arraybackslash}p{#1}}
\maketitle

\abstract{In recent years, Large Language Models (LLM) have demonstrated impressive capabilities in the field of natural language processing (NLP). This paper explores the application of LLMs in negotiation transcript analysis by the Vanderbilt AI Negotiation Lab. Starting in September 2022, we applied multiple strategies using LLMs from zero shot learning to fine tuning models to in-context learning). The final strategy we developed is explained, along with how to access and use the model.  This study provides a sense of both the opportunities and roadblocks for the implementation of LLMs in real life applications and offers a model for how LLMs can be applied to coding in other fields.}

\section*{Introduction}
\hspace{0.5cm}The field of natural language processing (NLP) has taken the spotlight with significant advancements in recent years, largely due to the emergence of Large Language Models (LLMs) and ChatGPT. These models are pre-trained on diverse and large datasets and demonstrate remarkable capabilities in understanding and generating human language. As LLMs continue to evolve and improve, new possibilities and capabilities for application across multiple domains are constantly emerging. Our study tests whether we can apply LLMs to automatically classify each sentence (or thought unit or speaking turn) in a negotiation transcript based on categories relevant to negotiation scholars (e.g. Weingart et al., 1990\cite{weigart1990}; Gunia et al., 2011\cite{gunia}; Adair \& Brett, 2005\cite{brett2014}).

One area where LLMs excel is in classification tasks, such as sentiment analysis, document classification, and summarization (Naveen et al., 2023).  LLMs are good at classification tasks because of contextual understanding, transfer learning, and semantic feature extraction (Kalyan, 2024\cite{Kalyan2024}). This capability can be used to automate the slow and costly process of coding transcripts, which is a procedure used in many negotiation studies.  Scholars in many fields (e.g., psychology, management, or communications) study negotiations by having people conduct simulated negotiations, audio recording them, and looking for patterns in negotiators’ verbal behaviors.  For example, scholars may be interested in information sharing in negotiations and when it is most effective (Adair et al., 2001\cite{adair2001}), the impact of cooperative orientation on use of threats (Weingart, et al., 1993\cite{weigart1993}), or the impact of supporting arguments on negotiation outcomes (Putnam \& Jones, 1982\cite{put1982}).  A necessary step in these kinds of studies is to develop a coding scheme, train coders, ensure inter-coder agreement, and have those coders code hundreds or thousands of speech units. This process of human coding can take months, and cost thousands of dollars. 

Researchers can save time and money by automating this coding process. Additionally, automated coding reduces the risks that are inherent in human coding – such as coder fatigue due to coding too many transcripts in a sitting or coder drift in applying codes across a number of transcripts.  Indeed, having an automatic coding process may make some projects feasible that previously were not feasible. This study explores whether, and how, LLMs can be used to replace (or supplement) human coders, which should increase both the efficiency and reliability of coding for negotiation research.  What we learned in the process of negotiation coding may also be applicable to other coding-heavy research in areas such as medical summarization of patient visit notes, analyzing corporate financial reports, or any number of similar tasks. 

To do this research Ray Friedman from the Owen School of Business at Vanderbilt and Jesse Spencer-Smith and Jaewoo Cho from the Vanderbilt Data Science Institute set up the AI Negotiation Lab at Vanderbilt University (\url{https://AINegotiationLab-Vanderbilt.com/}) in September 2022. The mission of the AI Negotiation Lab is the application of LLMs to support the research on conflict management and negotiation.  This is the first project of this lab. The project is supported by Negotiation and Team Resources (\url{https://www.negotiationandteamresources.com/}). This paper provides a record of the multiple experiments we conducted, each informed by the success or failure of the prior experiments, as we developed the LLM Vanderbilt Negotiation Coder.  

\section*{Selecting a Coding Scheme}
\hspace{0.5cm}As a starting point, we had to choose a coding scheme that the LLM model would learn.  Negotiation scholars have used many different coding schemes.  Just as this project began, a group of scholars reviewed the published negotiation coding schemes and developed an omnibus or “master” coding scheme which we call the Jäckel  Master (Jäckel  et al., 2022\cite{jackel}).  The Jäckel  Master has 47 codes, with definitions of each and example sentences to be used in the training of coders.  Three of those codes could not be used in this project since coding them requires knowledge of specific elements of the negotiation scenario. For example, we could not code “lying” since the model would have no basis to judge what was true and false. The two other codes not used were “omission” and “use of extreme anchors”. We also dropped “change of mode” since submitted transcripts may not be set up to indicate when, for example, negotiators switched from live to online negotiations.  We initially attempted to build an LLM model that would use all 44 remaining codes, but this model would not even run. We quickly learned that LLM categorization typically does not work well with more than about ten to twenty codes. At this point, we worked with Elizabeth Jäckel  to reduce the number of codes, by selecting the most frequently used codes (based on her transcript samples) and by combining some codes. The result was 18 codes plus an “other” category, producing 19 codes.  The initial full Jäckel  coding scheme is shown in Table \ref{tabel1}, and the simplified 19-code version is shown in Table \ref{tabel2}. 

\begin{table}[H]
\centering
\caption{Simplified Jäckel  et al (2022) Coding Scheme}
\begin{tabular}{|l|l|}
\hline
1 & \textbf{Providing positional information} \\
2 & \textbf{Asking positional information} \\
3 & \textbf{Providing priority-related information} \\
4 & \textbf{Asking for priority-related information} \\
5 & \textbf{Providing preference-related information} \\
6 & \textbf{Asking for preference-related information} \\
7 & \textbf{Clarification} \\
8 & \textbf{Single-issue activity} \\
9 & \textbf{Multi-issue activity} \\
10 & \textbf{Rejecting Offer} \\
11 & \textbf{Accepting Offer} \\
12 &\textbf{Contentious Communication} \textit{(Stressing Power, Criticism, Threat, Hostility)} \\
13&\textbf{Substantiation} \textit{(Substantiation, Asking for substantiation and Rejecting substantiation)} \\
14&\textbf{Positive Statements} \textit{(Positive affective reaction and Positive relationship remarks)} \\
15&\textbf{Negative Statements} \textit{(Negative affective reaction and Negative relationship remarks)} \\
16&\textbf{Humor} \\
17&\textbf{Active listening} \\
18&\textbf{Procedural comments} \textit{(Procedural suggestion, Procedural discussion, Time management)} \\
19&\textbf{Other} \\
\hline
\end{tabular}
\label{tabel2}
\end{table}

\begin{table}[H]
	\centering
	\caption{Overview of Categories and Codes}
	\scriptsize
	\renewcommand{\arraystretch}{2}
	\begin{tabular}{L{1.5cm} L{1.5cm} L{1.5cm} L{1.5cm} L{1.5cm} L{1.5cm} L{1.5cm}}
	\toprule
	\raggedright{\textbf{Acts of providing and asking about negotiation-related information}} & \raggedright{\textbf{Offers}} & \raggedright{\textbf{Acts of persuasive communication}} & \raggedright{\textbf{Socio-emotional statements}} & \raggedright{\textbf{Unethical behaviors}} & \raggedright{\textbf{Acts of process-related communication}} & \textbf{Residual Category} \\
	\midrule
	Providing priority-related information & Single-issue activity & Substantiation & Negative affective reaction & Omission\textsuperscript{a} & Procedural suggestion & Interruption of the conversation \\
Asking for priority-related information & Multi-issue activity & Asking for substantiation & Positive affective reaction & Threat & Procedural discussion & Inaction \\
Providing preference-related information & Requesting action & Stressing power & Active listening & Lying\textsuperscript{a} & \raggedright Time management & Others \\
Asking for preference-related information & Requesting for offer modification & Rejecting substantiation & Humor & Hostility & \raggedright Change of mode & \\
Asking for positional information & Rejecting offer & Interrupting & Positive relationship remark & \raggedright Use of extreme anchors\textsuperscript{a} & & \\
Providing positional information & Accepting offer & Criticism & Negative relationship & & & \\
Facts/Additional information & & Encouragement & Personal communication & & & \\
Extension questions & & Positional commitments & Nonpersonal chit-chat & & & \\
Additional issues & & Avoiding & Future-related communication & & & \\
Clarification & & & Apologizing & & & \\
\bottomrule
		
	\end{tabular}
\hspace{0.5cm} \parbox{0.9\textwidth}{\textsuperscript{a} can only be coded with data from experimental settings where role instructions and information given to the negotiators are disclosed to coders.}
\label{tabel1}
\end{table}

\section*{Creating Ideal Sentences}
\hspace{0.5cm}Our next step was to develop a list of examples sentences for each code that would be extensive enough for the model to learn the codes and test applying them.  We call this the “Ideal Sentences” approach since we use sentences that were deemed perfect exemplars of each code.  Some example sentences were provided in the coding manual for Jäckel  et al (2022\cite{jackel}), more examples were generated by Elizabeth Jäckel , and we asked ChatGPT to provide even more examples (which were checked by Elizabeth Jäckel  to make sure that they were indeed accurate examples of that code). At this point, we had 30-90 sample sentences for each of the 19 codes (we needed fewer for more easily understood codes like “offer rejected”, but more sample sentences for more complicated ideas like “asking for priority-related information”).  We ended up with 770 sample sentences.  Table \ref{table3} shows examples of sample sentences that were created.

\begin{table}[H]
\centering
\caption{Sample Sentences of Jäckel  Simplified Coding Scheme}
\begin{tabular}{|p{6cm}|p{6.5cm}|}
\hline
\textbf{Code} & \textbf{Generated Sentence} \\
\hline
Asking positional information & What is the maximum price you can pay for these materials? \\
\hline
Providing priority-related information & We prioritize growth and expansion over maintaining a conservative business model. \\
\hline
Single-issue activity & Would you accept a payment of \$5000 now and \$5000 upon completion? \\
\hline
Substantiation & We can't afford to overlook the potential consequences of this decision. \\
\hline
Rejecting Offer & I'm sorry, but this just isn't going to work for us. \\
\hline
Accepting Offer & That’s very generous of you. I accept. \\
\hline
\end{tabular}
\label{table3}
\end{table}

\section*{Three LLM Model Coding Strategies}
\hspace{0.5cm}We tested three coding strategies with these 770 sentences: zero-shot, fine-tuning a new model, and in-context learning. In each case, the model was deemed accurate for cases where it assigned a code to a sentence that the sentence was formulated to represent.

\textit{Zero-Shot.} For Experiment 1, we implemented a zero-shot strategy, which refers to using machine learning models “out of the box” – that is, just as they exist, without any domain-specific adaption (Xian, et al, 2020). In this strategy we tested whether established models (such as GPT3 or BART or BERT) can successfully code negotiation sentences based just on preexisting general knowledge about language that is built into those base models.  We tried this approach using open-source models BERT (Bidirectional Encoder Representations Transformers) from Google and BART (Bidirectional and Auto-Regressive Transformers) from Meta.  We asked these models to code the 770 sample sentences that we created.  This was done using their established “Inference API” where you put text in one window (in our case the sentences to be coded) and the “class names” in another window (in our case the codes), and the model allocates the material to the categories. It was clear that some codes could be understood based on general language knowledge (such as “Negative Statements”) while others made no sense to the LLM based on its general language knowledge. For example, a code that gave the LLM model trouble was “asking for preference related information”.  This zero-shot strategy accurately coded only about 20\% of the sentences we tested (the results were the same for BERT and BART).  The poor performance was expected since the models had only learned general language, and not the specialized language of negotiation scholars. Still, this was an important first step to confirm that domain adaptation was needed. It also provided a baseline to see how much improvement we could make as we moved to fine-tuned models.  

\textit{Fine-Tuning.} For Experiment 2, we implemented a fine-tuning strategy. Fine-tuning is when you train an existing model with your own domain-specific data to add specialized knowledge in a specific domain. This produces a new LLM model (the base model + domain specific information), separate and distinct from initial base model. In the AI and data science literature, fine-tuning is used to do various tasks like classification, sentiment analysis, and topic classification of news articles (e.g. are they finance, sports, or tech articles) (Lv et al., 2023\cite{lv2023}). 
	
	We fine-tuned the BERT model with information from the Jäckel  coding scheme, including definitions of the 19 codes and 527 of the 770 sample sentences we had written, creating a new domain-adapted BERT model that we called BERT-NegCodingJäckel . We chose BERT versus models like GPT4 as the base model for two reasons.  First, its relatively small size meant that the model could be loaded onto Hugging Face and directly accessed by negotiation scholars for free.  Second, at the time (spring of 2023) the larger LLMs (like GPT4) did not permit API access and only allowed users to access them in a ChatGPT-like chatbot interface since those models were carefully protected.  After creating our BERT-NegCodingJäckel  domain-adapted model we asked the model to code the 243 sentences examples that were not used in the fine-tuning.  Since we knew the code each sample sentence was designed to represent, we could evaluate how often the model assigned that code.  This approach produced accurate results for 61\% of the sentences, which was a major improvement from zero-shot. 
	
\textit{In-Context-Learning.} For Experiment 3, we implemented in-context learning.  In-context learning refers to providing the base model with one set of instructions (a prompt) that combines the training material with the analysis instructions. This makes the training material more “top-of-mind” for the model when doing the analysis.  An additional benefit is that in-content learning allows the use of much bigger and more powerful base model. Recall in fine tuning we added domain specific knowledge to the base model and with the idea that we would make the entire new model available for researchers. Having to make the base model with domain specific knowledge available to users meant we were limited to using a smaller model. In contrast, with in-context learning we only needed to make our prompt available to the researchers along with access to the publicly accessible base LLM model.  This allowed us to use a larger and more powerful base model. The downsides of in-context learning are a) each analysis may produce somewhat different results, since the model is learning anew (and slightly differently) with each analysis, and b) since all the training information needs to be contained in each prompt, the prompt needs to be large and the model has to allow such a large prompt. That was not possible with programs like BERT, which had a limit of 512 tokens of input at the time (tokens are units like letters, numbers, or symbols for machines).  In-context learning only became practical in the summer of 2023, when new models, like Claude100k (later called Claude1 after newer versions were added) and GPT4 were released.  Claude100k allowed 100,000 tokens of input.  For more details on the quantity of input allowed in different LLMs, see Appendix I. 
	
	We tried in-context learning with both GPT4 and Claude100k, with a prompt that included: the code definitions and about 70\% of the sample sentences (selected at random).  Claude100k had better results (93\%) than GPT4 (86\%), both of which were far better than domain-adapting the BERT model. Based on these results we committed to in-context learning. Furthermore, since Claude100k gave better results than GPT4, we focused on Claude. 
	
\textit{Output Format Consistency.} We provided instructions in the prompt to have the output include, in order: the sentence number, the sentence that was coded, and the assigned code, with certain punctuation. But sometimes the model would change the order and punctuation, which created problems producing readable reports. This inconsistency of output format can happen in LLMs due to the stochastic nature of LLM generation.  LLMs are stochastic because LLMs “think” through responses to prompts, and it’s thinking process can vary.  To understand this intuitively, if an LLM is asked to find the last word in the sentence “In the negotiation, we were unable to reach \_\_ ” the model (like humans) would know that words like “terms” or “agreement” would fit, but not words but not words like “gorilla”. Understanding language leads the model to exclude “gorilla” but it might reasonably choose “terms” one time and “agreement” another time. The same idea applies to output format.  

In addition, some variance in formatting was introduced because at the time we only had access to the models through chatbot interfaces (like ChatGPT and Poe).  Those interfaces are built to make interacting with the model be like a conversation, but in doing so they provide the user with far less control over how the model is given instructions. So, there was a limit to how specific our instructions-–including formatting--could be. These issues were resolved through various prompting techniques and by gaining access to the direct Application Programing Interface (API) for Claude100k through Amazon AWS Bedrock, which allowed control over parameters like “temperature”. Temperature refers to the “creativity” of the LLMs and can be set to reduce variance in output.  But as we added more content to the prompts to solve the output problem, we began to push the input limits of Claude1.

\begin{figure}[H]
	\centering
	\includegraphics[width=\textwidth]{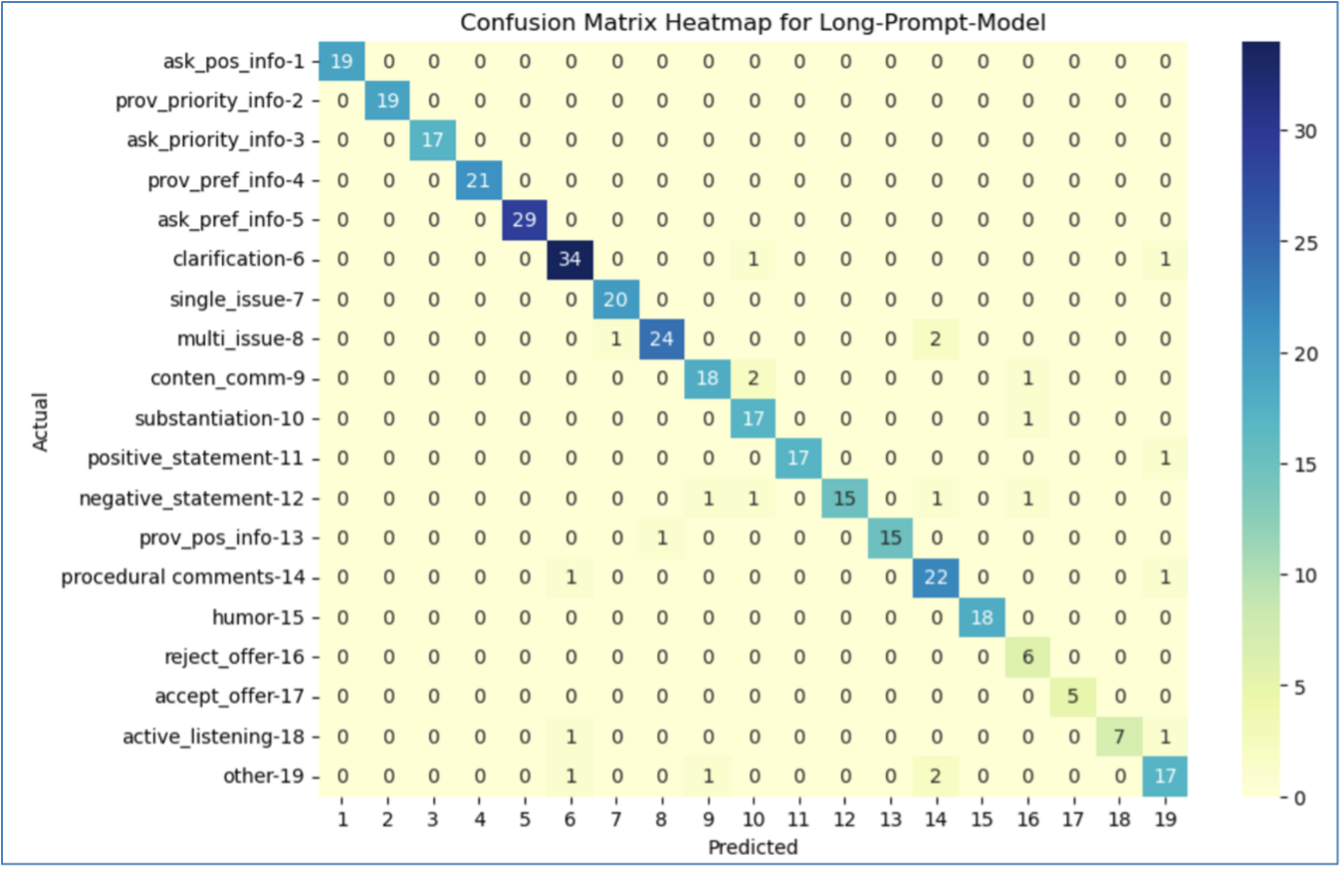}
	\caption{Confusion Matrix for Claude2 In-Context Learning of “Ideal Sentences” }
	\label{conf1}
\end{figure}

Fortunately, Claude2 was introduced in the fall of 2023, which allowed us to write even longer prompts that mostly solved the output format problems and gave us the room to add additional prompt modifications. With these adjustments using Claude2 the level of accuracy in coding sentences was 96\%.  Table \ref{table4} compares results for our three coding experiments, and Figure \ref{conf1} shows the best results achieved using Claude 2.

\begin{table}[h!]
\centering
\renewcommand{\arraystretch}{1.5}
\caption{Compare Coding Strategies for “Ideal Sentence” Approach}
\begin{tabular}{|L{2cm}|L{2cm}|L{3cm}|L{2.5cm}|L{2.5cm}|L{1.5cm}|}
\hline
\textbf{LLM Model} & \textbf{Method} & \textbf{Coding Scheme} & \textbf{Training Data} & \textbf{Testing Data} & \textbf{Accuracy} \\
\hline
BERT\textsuperscript{*} & Zero Shot & Jäckel  19 Categories & Ideal Sentences & Ideal Sentences & \~20\% \\
\hline
BART\textsuperscript{*} & Zero Shot & Jäckel  19 Categories & Ideal Sentences & Ideal Sentences & \~20\% \\
\hline
BERT\textsuperscript{*} & Fine-tune & Jäckel  19 Categories & Ideal Sentences & Ideal Sentences & 61\% \\
\hline
GPT4 & In-context & Jäckel  19 Categories & Ideal Sentences & Ideal Sentences & 86\% \\
\hline
Claude 1 & In-context & Jäckel  19 Categories & Ideal Sentences & Ideal Sentences & 93\% \\
\hline
Claude 2 & In-context & Jäckel  19 Categories & Ideal Sentences & Ideal Sentences & 96\% \\
\hline
\end{tabular}
\begin{flushleft}
\textsuperscript{*}\textit{We used the BERT base model and the BART “bart-large-mnli” model.}
\end{flushleft}
\label{table4}
\end{table}

\section*{Transition to a Transcript-Based Training Approach}
\hspace{0.5cm}After all our refinements, we reached an accuracy level of 96\%, which was extremely strong.   We thought we might have our final model.  As a final check, we applied this model to four complete full negotiation transcripts that were coded for this project using the Jäckel  19 code scheme.  The level of human-model match dropped from 96\% to 30\%.  What became clear is that we had taught the model with ideal sentences that were clear examples of codes, while real-life conversations are more halting, broken, incomplete, and ambiguous.  Moreover, the humans who coded those five transcripts were able to judge a sentence based on the broader context – the flow of the conversation, not just the sentence.  

	It became clear that we needed to train the model, not just with a set of “pure” example sentences, but also with sets of full, realistic, transcripts that had already been human-coded.  Since the Jäckel  et al (2022)\cite{jackel} coding system was new, there was no corpus of transcripts. that were coded using their scheme.  We turned to negotiation studies that had large numbers of human-coded transcripts. Here the problem was that each study used a coding scheme, chosen for that study’s particular research questions and even when different researchers used the same coding scheme, how the same set of codes was applied could vary across studies, based on differences in conceptual focus or coder tendencies. 
	
It became clear that to validate the transcript-trained LLM model against human coders we needed to a) train the model with the \textit{specific codes} used in that set of transcripts, and b) accept that what the model learns is the particular way that set of codes was used by that set of coders.  For this approach the in-context model prompt for each coding scheme would need to include that study’s code definitions, human-coded transcripts for the model to learn from (we ended up including five, after testing more transcripts and fewer transcripts), and several other elements (discussed below).  It also had to include the transcripts that we were asking it to code.

\textit{From One-to-Many Coding Schemes.} This approach moved us away from thinking that our LLM model would provide a single coding scheme and towards providing future users with several different options for coding. Once we had a method to train a model to code using one coding scheme, it should be relatively easy to apply that method to develop and validate models using other coding schemes.  We went from planning to offer “the” coding scheme, to planning to offer an array of coding schemes.  Scholars would be given information about each coding scheme, and assess which they found relevant for their purposes, and apply that coding scheme to their own transcripts. 

\textit{Unit of Analysis.} The move from ideal sentences to full transcripts required us to think about the unit of analysis.  In the ideal sentences, the unit was the sentence.  That was easy to do since we produced the sentences.  Moving to full transcripts there are three units of analysis that could be used.  First, is the speaking turn – the content that is spoken before another person starts speaking.  This could be a word (e.g. “yes”), a sentence, or multiple sentences.  Second is the sentence, defined as what is spoken prior to one of these three types of punctuation: period, question mark or exclamation point.  Third, is a thought unit, which could be a sentence, or part of a sentence (a long sentence might have several thought units).  Speaking turns are easy to identify but can produce ambiguity about what to code, especially if the speaking turn has multiple sentences.  A thought unit is easier to code, but it takes almost as much time to have humans create the thought units reliably as it does to have them do the actual coding.  Sentences are easier to identify than thought units and easier to code than speaking turns, especially when speaking turns are long, but there is some randomness about what a given transcriber decides should be a period (ending the sentence) or a comma (continuing the sentence).  In the end, we realized that for model learning, we had to use the unit of analysis had been used by the scholars who provided the code and the coded transcripts for validation. We leave it to the user to decide their unit of analysis. \footnote{Even a thought unit can fit into more than one coding category in some instances. Researchers use dominance schemes to address this problem. For example, if researchers are coding at the speaking turn level, they might allow up to three codes per thought unit and then chose the code to use to analyze their data based on the theory and research questions they are testing.  Another approach is to build a dominance scheme into the code itself. In this option, the researchers tell the coders which code dominates when two or more could apply (Kern et al., 2020\cite{Kern}). We have not yet built dominance schemes into our LLM models.}

\textit{Infrequent codes.} By moving from ideal sentence to real conversations, we could no longer ensure that each code was represented with sufficient examples in the learning material.  An example of an infrequent code might be “offer reject” since in many transcripts there may not be any cases of explicit rejection, and if there is a rejection, it might occur only once towards the end. By contrast, codes like “information” would occur frequently.  For codes that did not have at least 15 examples in the learning transcripts, we created sentences that would fit the code and added them to the prompt. \footnote{We consulted with Johnathan Gratch, Professor of Computer Science and psychology at USC, about the issue of low base-rate codes.  He reported that Krippendorff (1980)\cite{KK1980} -- whose work has been adopted in computational linguistics (Calretta, 1996)\cite{Carletta1996} -- explains that you need five naturally occurring examples of a code in a developmental corpus to be able to have a reliable code for human coding, but there has not yet been any guidance for LLM coding. }

\textit{Validation.}  Given that we were no longer dealing with ideal examples, we would not expect anything close to the 96\% match we achieved with ideal sentences in Experiment 3 with in-context 
learning but were hoping to achieve an acceptable level of agreement between the model and the human coders. We call this “Validation Step 1”. This step ensures that the model can sufficiently replicate human coding for a given study.  However, in Validation Step 1, both the training transcripts and the testing transcripts use the same negotiation simulation, so the specific topics discussed are the same. However, users of the LLM model are not likely to all use the same simulation.  We needed to ensure that the model can properly code transcripts of negotiations using different simulations. “Validation Step 2” tests the model’s success with transcripts from different negotiation simulations.  We expect that model-human matches will be harder to achieve in Validation Step 2 than Validation Step 1, but we hope that the match level remains acceptable.

For both Validation Steps 1 and 2, we refer to model “matching” with the human coders, not the model “accuracy”.  In the ideal sentence approach, each sentence was formulated to be a perfect match with that code, so we could reasonably say that a higher level of model match with that assigned code meant it was “accurate”. With the transcript training approach, however, there was much more room for human coders to be wrong, in part due to human error and fatigue and drift, but also because they were coding real, more ambiguous, sentences.  Human coders might themselves have differences of opinion about which code applied.  So, rather than call it “accurate” when the model code matched the human code, we simply refer to it as a “match”. 

What level of match is an acceptable is an unresolved issue. We provide more detailed statistics on interrater agreement for our first model in Friedman et al (2024), but there are no definitive break points for those statistics, and it is relatively easy to achieve momentarily high human-human kappas with repeated training and coding without necessarily achieving ongoing inter-rater alignment. Also, as just mentioned, close evaluation of past coding by humans reveals many cases where reasonable coders could disagree. We discuss below an approach we call “mismatch analysis” where we ask independent coders to evaluate whether the original human coders or the LLM coder assigned the right code. More broadly, we go through an extensive process of testing the data in ways that can help build confidence in our process and the results. 

\section*{Coding Consistency Measure}
\hspace{0.5cm}As we tested different coding strategies, we realized that the model might not always code a given sentence the same way every time, especially when using in-context learning, which has the model learn afresh each time it is applied to a transcript.  To test how consistently the model coded a transcript’s sentences, we decided to repeat the coding process five times, producing five codes for each sentence. This process was then built into our program – all coding done by the model would be repeated five times. Some variation across five runs might be acceptable, but before we would allow the model to report a code to users, the model had to give the same code at least three out of five times.  If the model could not achieve that level of consistency, the model would report that “no code was assigned”. While this results in scholars having to hand code that subset of sentences, the benefit is greater confidence in the sentences that the model did code consistently. The model reports consistency level: moderate (three out of five), high (four out of five) or perfect (five out of five) for all codes.

\begin{table}[H]
\centering
\renewcommand{\arraystretch}{1.5}
\caption{AI Coder Consistency Measure (using 5 runs)}
\begin{tabular}{|L{2cm}|L{2cm}|L{2cm}|L{2cm}|L{2cm}|}
\hline
\raggedright \# of runs that assign the same code: & \textless{} 3 out of 5 & 3 out of 5 & 4 out of 5 & 5 out of 5 \\
\hline
Label: & \raggedright Unreported. Consistency too low for model to report a code & Moderate Consistency & High Consistency & Perfect Consistency \\
\hline
\end{tabular}
\label{table5}
\end{table}

\section*{Technical Challenges}
\hspace{0.5cm}One problem we faced was that Claude 2 would skip some sentences, exhibiting what has come to be known as model “laziness”. Another problem was that the model would sometimes correct the grammar of the sentences it was coding, which might change the sentence meaning and make it hard to match with the original sentences.  We suspected that these model behaviors might be due to reaching the capacity limits of Claude 2, so we turned to Claude 3 Opus–Anthropic’s most powerful model as of spring, 2024.  Moving to Claude 3 Opus resolved these issues. 

\textit{Length of transcript to be coded.}  We tested how long a transcript could be that we asked it to code. Trial and error led to the realization that it could only handle well about 100 sentences.  We programed our code to calculate the sentence count of submitted transcripts and split that transcript into two (or more) segments which were then coded separately (but the results are shown together in the output).

\section*{First Model Developed}
\hspace{0.5cm}The first model we developed with this method was Model 1, using a 13-code coding scheme and transcript database provided by Aslani et al (2014)\cite{aslani2014}. A detailed report of this model, including the coding scheme, validation data, and how to submit transcripts for analysis, is reported in Friedman et al (2024)\cite{ray2024}.  The next sections describe several substantive and technical modifications that we tried before finalizing Model 1. Over time, additional models will be provided, each with an orientation document like Friedman et al., (2024)\cite{ray2024}. 

\section*{Improvement Experiments}
\hspace{0.5cm}\textit{Substantive.} As we went through testing, we learned that the model’s match level improved (had a higher match rate) with several additions. First, we instructed the model to pay attention to who was speaking (e.g., buyer or seller) and what was said in the sentences before and after speaking unit being coded.  This was important when using full transcripts to train and code but would not have been relevant in our earlier “ideal sentence” approach since those sentences were not within a conversation.  Second, we added more instructions about certain codes. That was done because, after looking at the confusion matrix of the initial runs, certain codes were being mixed up at especially high levels (the codes for “information” and “substantiation”). The model was improved by adding more detailed instructions for those codes. We also tested whether the model would improve by adding a paragraph describing the simulation case, including who the parties were and what the issues were. This addition, however, did not improve the model’s performance. See Friedman et al (2024)\cite{ray2024} for more details.

\textit{Technical.} To further improve the model’s performance, we tested incorporating XML tags (HTML tags) in the prompts and using chain of thought (asking the model to explain its coding choice in every case).  These strategies did not improve the human-model match level; in fact, they each reduced the match percentage by a few percentage points. Based on those findings, we dropped these ideas for model improvement. We also tested whether adding more transcripts to the training would help. We learned that having more than five training transcripts made the results worse, probably due to reaching close to the capacity limit of the model.  We also experimented with changing the set of five transcripts chosen for training.  For Model 1, we had 70 transcripts to choose from, which provided 17,259,390 possible combinations of 5 transcripts. That was clearly too much to test empirically, so we needed a rule to guide us. We decided that we would first select all combinations of five which would include at least once case of all codes within the set of five. At that point, we randomly selected five combinations to test empirically. Among those, we used the one set that produced the highest level of human-model match.  Lastly, as Claude 3.5 Sonnet became available in June of 2024, we tried using that model, but the match level was just a few percentage points lower.  So, we stayed with Claude 3.0 Opus. 

\section*{Model 1 Results}
\begin{figure}[H]
	\centering
	\includegraphics[width=\textwidth]{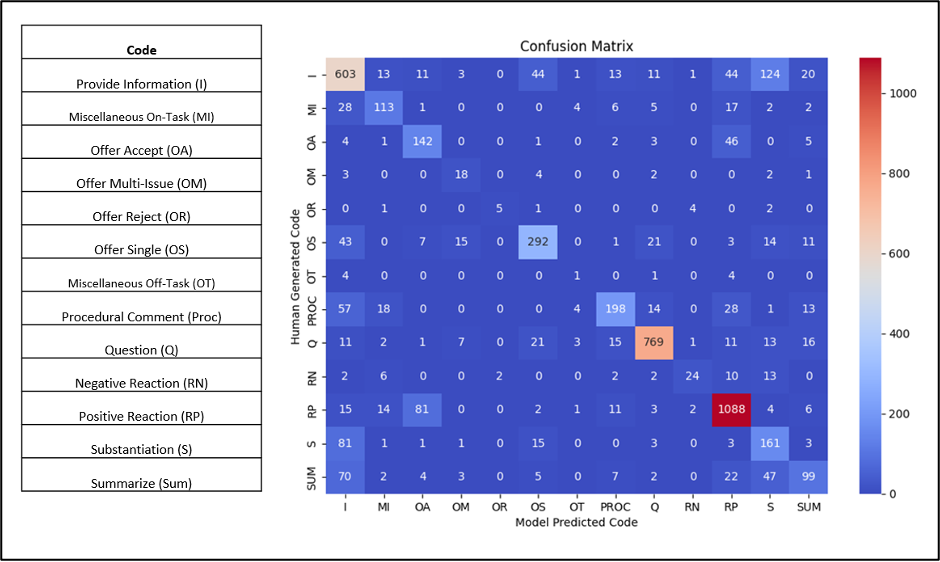}
	\caption{Confusion Matrix for Claude3 Model, Applied to Transcripts Using Sweet Shop Simulation}
	\label{conf2}
\end{figure}

\hspace{0.5cm}Here we provide summary results for Model 1.  The effectiveness of the model coding real conversations (transcripts, rather than ideal sentences) improved dramatically when we moved from training with ideal sentences to training with transcripts (from 30\% up to 70\%, and then 73\% with additional adjustments and a move to Claude3). See Tables \ref{table6} and Figure \ref{conf2}. We also conducted a mismatch analysis. This was done by training new coders and giving them 100 randomly selected cases of mismatch.  They were provided with a spreadsheet with the speaking turn (along with the two prior speaking turns to understand the context), and the codes assigned by the model and the human coders (shown without indicators of which code was the model code and which was the human code, and with the order varied to eliminate order effects).  These coders were asked to decide which of the two codes they thought was correct. This analysis revealed that the new coders agreed with the model’s coding choice (rather than the human-coders’ choice) 68\% of the time, suggesting a true accuracy level 91\% for the model.

However, this test had the advantage that the training and training transcripts both used the same negotiation simulation – \textit{The Sweet Shop}.  We further tested the model by asking it to code transcripts from a different simulation (6 from the \textit{Cartoon} negotiation, and 6 from the \textit{Les Flores} negotiation \footnote{The Les Florets transcripts were negotiations provided by Lowenstein and Brett (2007). The Cartoon transcripts were negotiations provided by Jeanne Brett (unpublished).  Samples from both studies were US.  The Sweet Shop exercise is available from \url{https://new.negotiationexercises.com/}.  Both Les Florets and Cartoon are available from \url{negotiationandteamresources.com}}). The match level came down to 65\%, which is still strong given that this level of agreement was reached for a 13-code coding scheme. Further analysis of mismatches suggested that approximately 82\% of the model codes were accurate. Full results are included in Friedman et al (2024)\cite{ray2024}, including kappa calculations, match levels by code, frequency of codes, details on the mismatch analysis, and a deeper assessment of the codes that were confused most often.  

\begin{table}[h!]
\renewcommand{\arraystretch}{1.5}
\centering
\caption{Compare Coding Strategies Tested with Full Transcripts (Real Conversations)}
\begin{tabular}{|L{1.5cm}|L{1.7cm}|L{2cm}|L{3cm}|L{3cm}|L{3cm}|}
\hline
\raggedright\textbf{LLM Model} & \textbf{Method} & \textbf{Coding Scheme} & \textbf{Training Data} & \textbf{Testing Data} & \textbf{Human-Model Match} \\
\hline
\raggedright Claude2 & In-context & Jäckel  19 Categories & Ideal Sentences & \raggedright 4 Human-Coded Transcripts & 30\% \\
\hline
Claude2 (Opus) & In-context & Aslani et al. & \raggedright 5 Sample Human-Coded Transcripts (from Sweet Shop Simulation) & \raggedright 70 Human-Coded Transcripts (from Sweet Shop Simulation) & 70\% (model accuracy estimate: 91\%) \\
\hline
Claude3 (Opus) & In-context & Aslani et al. & 5 Sample Human-Coded Transcripts (from Sweet Shop Simulation) & 70 Human-Coded Transcripts (from Sweet Shop Simulation) & 73\% (model accuracy estimate: 91\%) \\
\hline
Claude3 (Opus) & In-context & Aslani et al. & 5 Sample Human-Coded Transcripts (from Sweet Shop Simulation) & 12 Human-Coded Transcripts (6 from Cartoon Simulation and 6 from Towers simulation) & 65\% (model accuracy estimate: 82\%) \\
\hline
\end{tabular}
\label{table6}
\end{table}

\textit{Note on Language.}  We also tested if the model would code Chinese text, not just English.  We translated 27 randomly selected speaking units into Chinese using Google translate and asked the model to code them. The model produced the same code when the speech was in English as in Chinese in 24/27 cases.  In the few cases of a mismatch, the Google translation had a hard time conveying the idea in Chinese since the English was a colloquial phrase.  This suggests that the model can apply the codes to any language that it understands, and most major languages should be well-understood by the model. 

\section*{Discussion}
\hspace{0.5cm}This paper outlines the development of a customized LLM model for the purpose of coding negotiation transcripts. However, the steps and procedures we took provide a roadmap for applying LLMs to coding for social science research more generally. There are currently many scholars looking for ways to automate the time-consuming and expensive process of coding.  If AI assisted coding becomes possible, not only will it make current projects easier to complete but will also make it feasible to work with even larger data sets that would be beyond what humans could possibly code.

Our work suggests that in-context learning is likely to be the most effective way to code.  Zero-shot coding is not viable for any coding scheme that depends on large number of codes that identify special concepts, but it may still be possible if the concepts are ones often used in regular conversation (such as “stress” or “happiness”).  Domain-adapting a model also appears much less effective than in-context learning. Our work also suggests that while training with “ideal” examples is the cleanest way to code, it may not teach the model to understand the kind of language that is encountered in real conversation or speaking. The best approach for coding real conversations is to train on real conversations.

The challenge that creates is that you need a large corpus of coded material to train with – exactly when the purpose of using LLM coding is to avoid having to human code your data set. However, the number of transcripts you need to human code to build a LLM model should be much less than your whole data set.  For our model, it took five transcripts to provide in-context learning of the coding scheme, and we were able to validate the model using about 10-15 additional human-coded transcripts.  While we validated with 70 transcripts in our first validation step, we validated with 12 transcripts in our second validation step (although it may require more transcripts if the transcripts are shorter, or the coding scheme is more complicated). This suggests that you can validate a coding model by investing in human coding only a small subset of the total number of transcripts you need to code, leaving the bulk of the transcripts to be coded by the LLM. 

	We also learned to think differently about “accuracy”, realizing that human coded transcripts (even ones with two coders and higher inter-rater reliability) are likely to include many mistakes.  A model, unlike a human, does not get tired, or bored, or distracted, or feel that they have to meet a coding goal per hour to get through it all. This led to the use of “mismatch” analyses between the model and humans in cases of human-model code mismatches. 
	
	And we moved over the course of the project from trying to create an LLM that would implement one master coding scheme—that developed by Jäckel , et al. (2022)—to providing a template for building a model for any number of coding schemes. We have so far completed one (from Alsani et al, 2016), but expect to provide many more. This approach puts less stress on gaining perfection for one model and allows the user to have much more control, since there will be several models to choose from. By having developed a strategy for building Model 1, we will be able to quickly build additional models for other codes. We will still try to build a model for the Jäckel  et al. (2022) codes once human-coded transcripts are available, but as one model among many. 
	
	There are myriad other lessons one might take from our experience, but the biggest one is that with a technology that is changing so fast, you need to be prepared for frequent twists and turns.  This study might well be titled: “Building on the shoulders of redundancy. The wandering exploits of a research team using technology that is changing constantly and nobody fully understands.” Be prepared to test and experiment, since just when you are facing an insurmountable roadblock, the next round of technology updates may well provide the solution. 
\newpage
\appendix
\section*{\centering Appendix}
\hspace{0.5cm}Taking a strategy of in-context learning for LLMs requires awareness of the capacity limits of different models.  This is referred to as the “context length”, which is the maximum number of tokens or words that the model could remember within one session. After the maximum number of tokens is exceeded, the LLM will forget everything that happened within the current session and return to a blank state.  Tokens refers to units that the model processes which could be letters, or clusters of letters, or words, or spaces.
Context length is counted in number of “tokens” allowed in the prompt (the input sent to the model which provides instructions), which can be expressed in terms of an approximate word length.  

\begin{table}[h!]
\centering
\renewcommand{\arraystretch}{1.5}
\begin{tabular}{|L{4cm}|L{3cm}|L{6.5cm}|}
\hline
\textbf{Model (release date)} & \textbf{Context Length} & \textbf{Approximate Context Word Length} \\
\hline
BERT (10/2018) & 512 Tokens & 400 Words \\
\hline
GPT4 (3/2023) & 128,000 Tokens & 96,000 Words \\
\hline
Claude 1 (4/2023) & 100,000 Tokens & 75,000 Words \\
\hline
Claude 2 (11/2023) & 200,000 Tokens & 150,000 Words \\
\hline
Claude 3 (4/2024) & 1,000,000 Tokens & 750,000 Words \\
\hline
\end{tabular}
\end{table}


\newpage
\begin{thebibliography}{99}
\bibitem{adair2005}
Adair, W. L., \& Brett, J. M. (2005). The negotiation dance: Time, culture, and behavioral sequences in negotiation. \textit{Organization Science}, 16, 33-51.

\bibitem{adair2001}
Adair, W. L., Okumura, T., \& Brett, J. M. (2001). Negotiation behavior when cultures collide: The United States and Japan. \textit{Journal of Applied Psychology}, 86(3), 371–385. \url{https://doi.org/10.1037/0021-9010.86.3.371}

\bibitem{aslani2014}
Aslani et al. (2014), Measuring negotiation strategy and predicting outcomes: Self-reports,
behavioral codes, and linguistic codes.  Presented at the annual conference of the International
Association for Conflict Management, Leiden, The Netherlands. (Link to paper is posted on Vanderbilt AI Negotiation Lab website.)

\bibitem{brett2014}
Brett, J.M. (2014). Negotiating Globally: How to Negotiate Deals, Resolve Disputes, and Make Decisions Across Cultural Boundaries, 3rd ed., Jossey-Bass, San Francisco, CA.

\bibitem{Carletta1996}
Carletta, J. (1996). Assessing agreement on classification tasks: The kappa statistic. \textit{Computational Linguistics}, 22(2):249–254.

\bibitem{ray2024}
Friedman, R., Brett, J., Cho, J., Zhan, X.,… Coding Negotiations with AI: Instructions and Validation for Coding Model 1. arXiv:submit/5703075[cs.AI] 1 Jul 2024. 

\bibitem{gunia}
Gunia, B. C., Brett, J. M., Nandkeolyar, A. K., \& Kamdar, D. (2011). Paying a price: Culture, trust, and negotiation consequences. \textit{Journal of Applied Psychology}, 96, 774-789.

\bibitem{jackel}
Jäckel , Elisabeth \& Zerres, Alfred \& Hemshorn de Sanchez, Clara \& Lehmann-Willenbrock, Nale \& Hüffmeier, Joachim. (2022). NegotiAct: Introducing a Comprehensive Coding Scheme to Capture Temporal Interaction Patterns in Negotiations. Group \& Organization Management. 49. 1-41. \url{https://doi.org/10.1177/10596011221132600}

\bibitem{Kalyan2024}
Kalyan, K. S. (2024). A survey of GPT-3 family large language models including ChatGPT and GPT-4, \textit{Natural Language Processing Journal}, Volume 6, 100048, ISSN 2949-7191, \url{https://doi.org/10.1016/j.nlp.2023.100048}

\bibitem{Kern}
Kern, M. C., J. M. Brett, L. R. Weingart, and C. S. Eck. (2020). The “fixed” pie perception and strategy in dyadic versus multiparty negotiations. Organizational Behavior and Human Decision Processes 157: 143– 158.

\bibitem{KK1980}
Krippendorff, K. (1980). Content analysis an introduction to its methodology. London: Sage.

\bibitem{Loe2007}
Loewenstein, J., \& Brett, J. M. (2007) Goal framing predicts strategy revision: When and why negotiators reach integrative agreements. Proceedings of the 29th Annual Conference of the Cognitive Science Society, Nashville, TN

\bibitem{lv2023}
Lv,K., Yang,Y., Liu,T., Gao,Q., Qiu, X. (2023). Full parameter fine-tuning for large language models with limited resources \url{https://arxiv.org/abs/2306.09782}

\bibitem{put1982}
Putnam, L. L., \& Jones, T. S. (1982). Reciprocity in negotiations: An analysis of bargaining interaction. \textit{Communication Monographs}, 49(3), 171–191. \url{https://doi.org/10.1080/03637758209376080}

\bibitem{nav2023}
Naveed, H., Khan, A.U., Qiu, S., Saqib, M., Anwar, S., Usman, M., Barnes, N. and Mian, A., 2023. A comprehensive overview of large language models. \textit{arXiv preprint} arXiv:2307.06435.

\bibitem{weigart1990}
Weingart, L. R., Thompson, L. L., Bazerman, M. H., \& Carroll, J. S. (1990). Tactical behavior and negotiation outcomes. \textit{International Journal of Conflict Management}, 1, 7-31.

\bibitem{xian2020}
Xian,Y., Lampert, C.H., Schiele,B., Akata,Z. (2020). Zero-shot learning: A comprehensive evaluation of the good, the bad and the ugly. \url{https://arxiv.org/pdf/1707.00600}

\bibitem{weigart1993}
Weingart, L. R., Bennett, R. J., \& Brett, J. M. (1993). The impact of consideration of issues and motivational orientation on group negotiation process and outcome. \textit{Journal of Applied Psychology}, 78(3), 504–517.

\end{thebibliography}
\end{document}